# On Languaging a Simulation Engine

Han Liu [a,b] *, Liantang Li [a,b]

[a] *SOlids inFormaTics AI-Laboratory (SOFT-AI-Lab), College of Polymer Science and Engineering, Sichuan University, Chengdu 610065, China*

[b] *AIMSOLID Research, Wuhan 430223, China*

*\* Corresponding author: Han Liu ([happylife@ucla.edu](happylife@ucla.edu))*

## Abstract

Language model intelligence is revolutionizing the way we program materials simulations. However, the diversity of simulation scenarios renders it challenging to precisely transform human language into a tailored simulator. Here, using three functionalized types of language model, we propose a language-to-simulation (Lang2Sim) framework that enables interactive navigation on languaging a simulation engine, by taking a scenario instance of water sorption in porous matrices. Unlike line-by-line coding of a target simulator, the language models interpret each simulator as an assembly of invariant tool function and its variant input–output pair. Lang2Sim enables the precise transform of textual description by functionalizing and sequentializing the language models of, respectively, rationalizing the tool categorization, customizing its input–output combinations, and distilling the simulator input into executable format. Importantly, depending on its functionalized type, each language model features a distinct processing of chat history to best balance its memory limit and information completeness, thus leveraging the model intelligence to unstructured nature of human request. Overall, this work establishes language model as an intelligent platform to unlock the era of languaging a simulation engine.

# 1. Introduction

Languaging a material simulation—viz., leveraging human language to target a material simulator (see Fig. 1a)—is crucial in advancing the intelligence of human-machine interaction in materials modeling [1–3]. However, this languaging process is generally challenged by the miscellaneousness of simulation scenarios, modeling toolkits, and computing quantities [4–7] (see Fig. 1b). As such, there remains lack of language model (LM) that enables the precise transform of human language into a tailored simulator [8,9].

Owing to its ability in long-text sequence mapping [10,11], transformer-architected large language model (LLM) offers an attractive opportunity in revolutionizing the intelligent era of programming from "machine language" to "natural language" paradigm [12,13]. A few recent LLMs have demonstrated their capacity to serve as programming copilot in *line-by-line* coding a tailored simulator by human language instructions [9,14]. Despite its unprecedented intelligence, LLM is intrinsically a giant regressor with billions of hyperparameters tuned by a giant training set of text sequence pairs [15,16], so that the level of model intelligence is greatly limited by the size of model hyperparameters [17,18], the scope of training set [19], and the quality of text labeling [20,21]. As a result, when targeting a material simulation, the present language models are prone to falling into the common "hallucination" dilemma [22,23], namely, an inaccurate transform from human language to simulation engine.

Here, relying on three functionalized types of LM—i.e., LM-agents that regulate action-outputs [24–26], we introduce a three-module hierarchical architecture to mitigate the issues of language-to-simulation (Lang2Sim) conversion facing a standalone LLM-based foundation model. Unlike programming-copilots dedicated in generating a delicate code of simulation [20], our Lang2Sim platform excels at activating a chain of LM-agent actions to navigate the transform of human language in a pre-delineated landscape of simulation engines, as delineated herein by the three-module global hierarchy and, inside each module, the agent-assembly local hierarchy. Indeed, we find that, in analogy to the search-path scheme of gradient-decent optimization [27–29], the strategy of "navigation-in-hierarchy" enables us to target an optimal simulator in excellent agreement with its textual description, as exemplified herein by languaging a water adsorption simulator via lattice density functional theory (LDFT) [30–32]. Importantly, when running on User-Interface (UI), the languaging process exhibits a satisfactory extent of intelligence to interact with UI-detected texts by selectively memorizing the navigation history. As future opportunities, we expect that the functionality of Lang2Sim platform would be empowered by stimulations of new developments in that direction, including (i) templatized accommodation of miscellaneous simulators [33–37], (ii) mutual integration of simulations and machine learning [38–41], and (iii) tabulated schedule of all-in-one model execution [42–45].

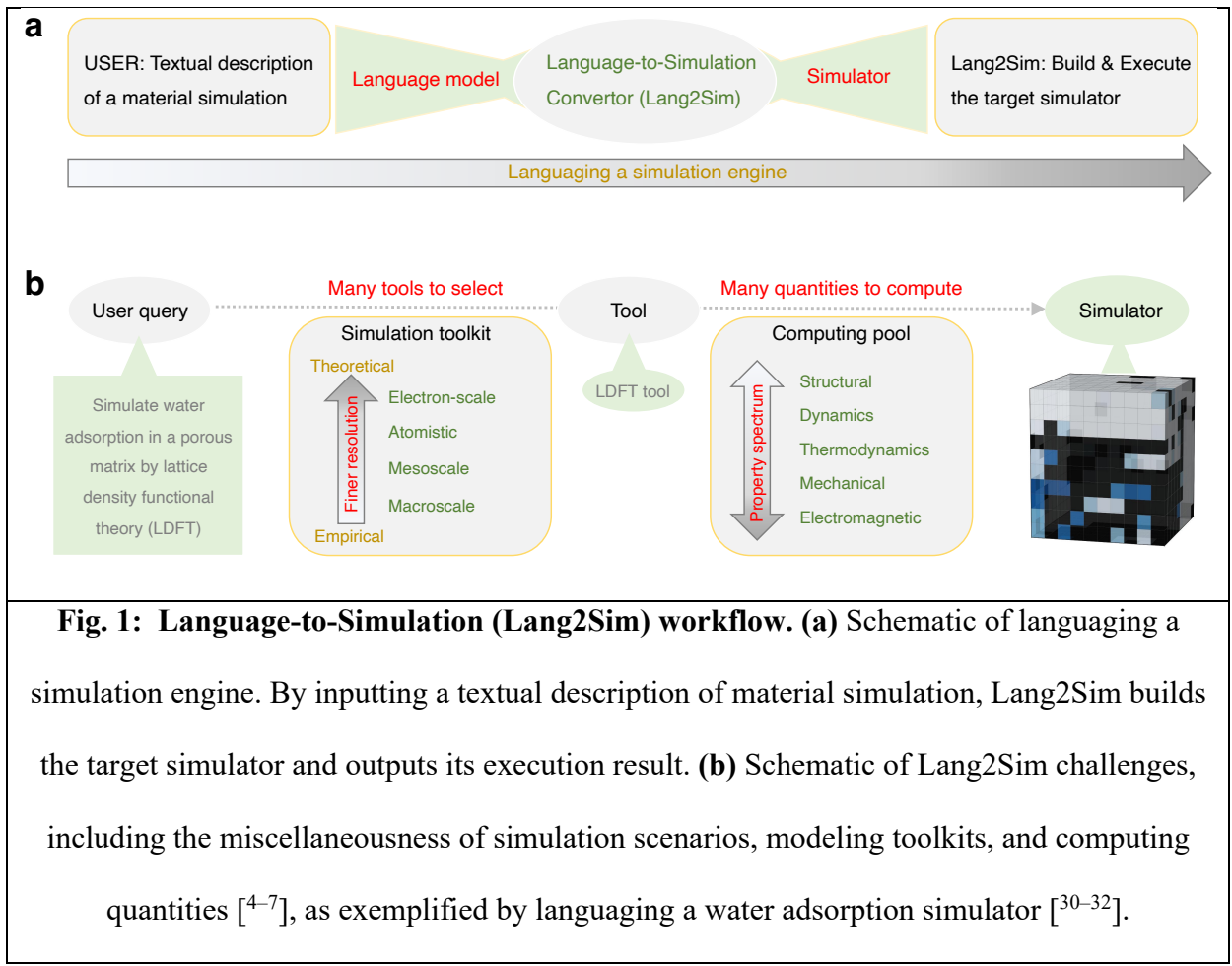

**Fig. 1: Language-to-Simulation (Lang2Sim) workflow. (a)** Schematic of languaging a simulation engine. By inputting a textual description of material simulation, Lang2Sim builds the target simulator and outputs its execution result. **(b)** Schematic of Lang2Sim challenges, including the miscellaneousness of simulation scenarios, modeling toolkits, and computing quantities [4–7], as exemplified by languaging a water adsorption simulator [30–32].

## 2. Results and Discussion

### 2.1 Lang2Sim platform built by three-module hierarchical architecture

To establish our conclusions, we now delve into the three-module hierarchical architecture that builds the Lang2Sim platform, as illustrated in Fig. 2, where each module is featured by a local hierarchy of LM-agent assembly. Figure 2a shows three types of LM-agents that are building blocks of the three modules, respectively, including LM-Type, LM-Sim, and LM-EXE, which are functionalized agents aiming to output an action of simulation type, simulator index, and simulator input, respectively [24–26] (see Sec. 2.2–2.4). To realize its functionality, each LM-agent has been carefully tuned to generate target actions by prompt-engineering and fine-tuning of a transformer-architected LLM [46,47] (i.e., an open-source LLaMA2 model herein [48]), or by a similarity search model using regular expression patterns that construct an output-formatted LM-EXE [49]. More details about the LM-agents are provided in the Methods section.

Built upon the three functionalized types of LM-agents, we then construct the Lang2Sim platform by assembling the agents of the same type into a monofunctional module and, subsequently, sequentializing these modules based on their functionality. Figure 2b illustrates the three-module sequentialization into a hierarchical architecture, which consists of (i) a Simulation-Type module built by LM-Type to rationalize the categorization of simulation types (see Sec. 2.2), (ii) a Simulator-Action module built by LM-Sim to customize the simulator input–output combinations (see Sec. 2.3), and (iii) a Simulator-Input module built by LM-EXE to transform textual input into simulator-executable input (see Sec. 2.4). More details about the Lang2Sim architecture are provided in the Methods section. From the general viewpoint of regression or classification model [7,33,34], by weaving the action-chain network among LM-agents, their hierarchical organization essentially depicts a clear "sim–lang" landscape of simulation engines as a function of human language (see Fig. 2b). Consequently, when inputting a textual description of material simulation, the landscape map plays the key role in navigating the conversion to its target simulator.

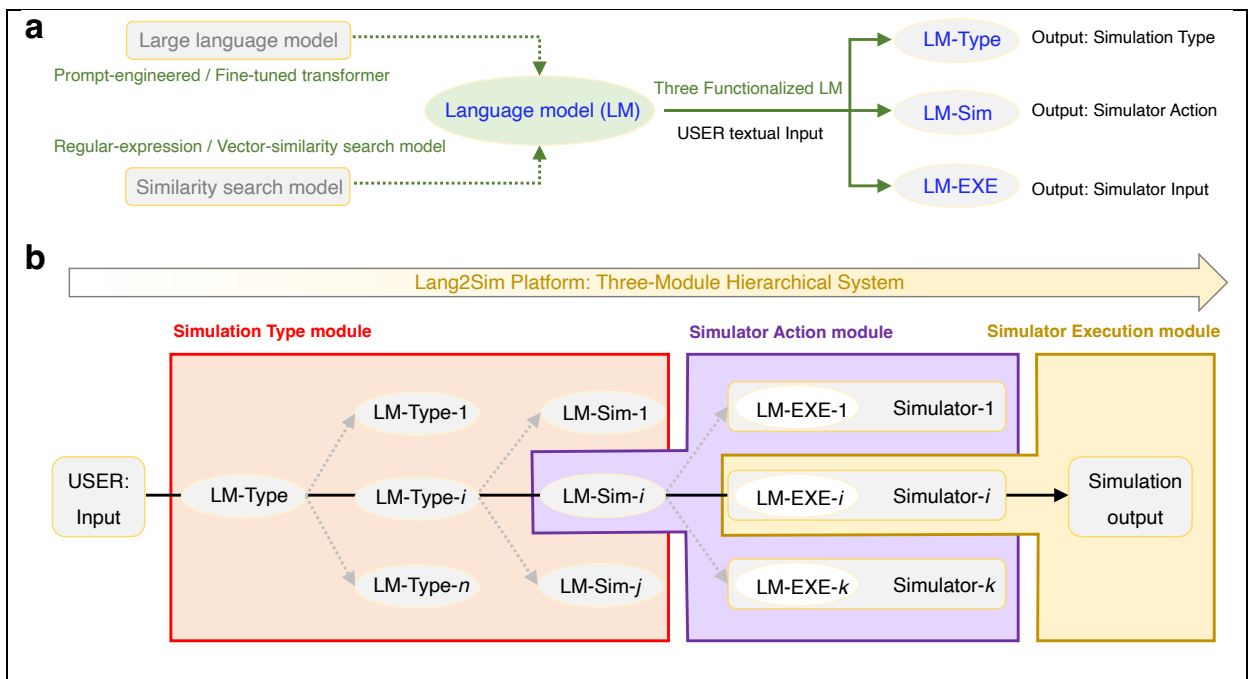

**Fig. 2: Architecture of Lang2Sim platform. (a)** Three functionalized types of language model (LM) built by large language model [47] and similarity search model [49]. When inputting a simulation text, the three models LM-Type, LM-Sim, and LM-EXE output the Simulation-Type, Simulator-Action, and Simulator-Input, respectively. **(b)** Three-module hierarchical architecture of Lang2Sim platform, including Simulation-Type, Simulator-Action, Simulator-Execution modules built by LM-Type, LM-Sim, and LM-EXE, respectively.

Figure 3 provides an example of the interplay between Lang2Sim modules and User-Interface (UI), by taking the example of languaging a water sorption simulation [30–32]. According to UI-detected text, one of the LM-agent chains on Lang2Sim platform (see Fig. 2b) is activated to target its chain-end simulator, i.e., a water adsorption simulator herein via 2D lattice density functional theory (2D-LDFT) []. The computation details of LDFT simulation can be found in the Methods section. It is worth mentioning that, despite its brevity, this "toy" interplay turns out to activate a series of underlying LM-agents within a complex sim–lang landscape (see Sec. 2.2–2.4), thus significant to revealing the platform intelligence in precise transform of human language to simulation engine.

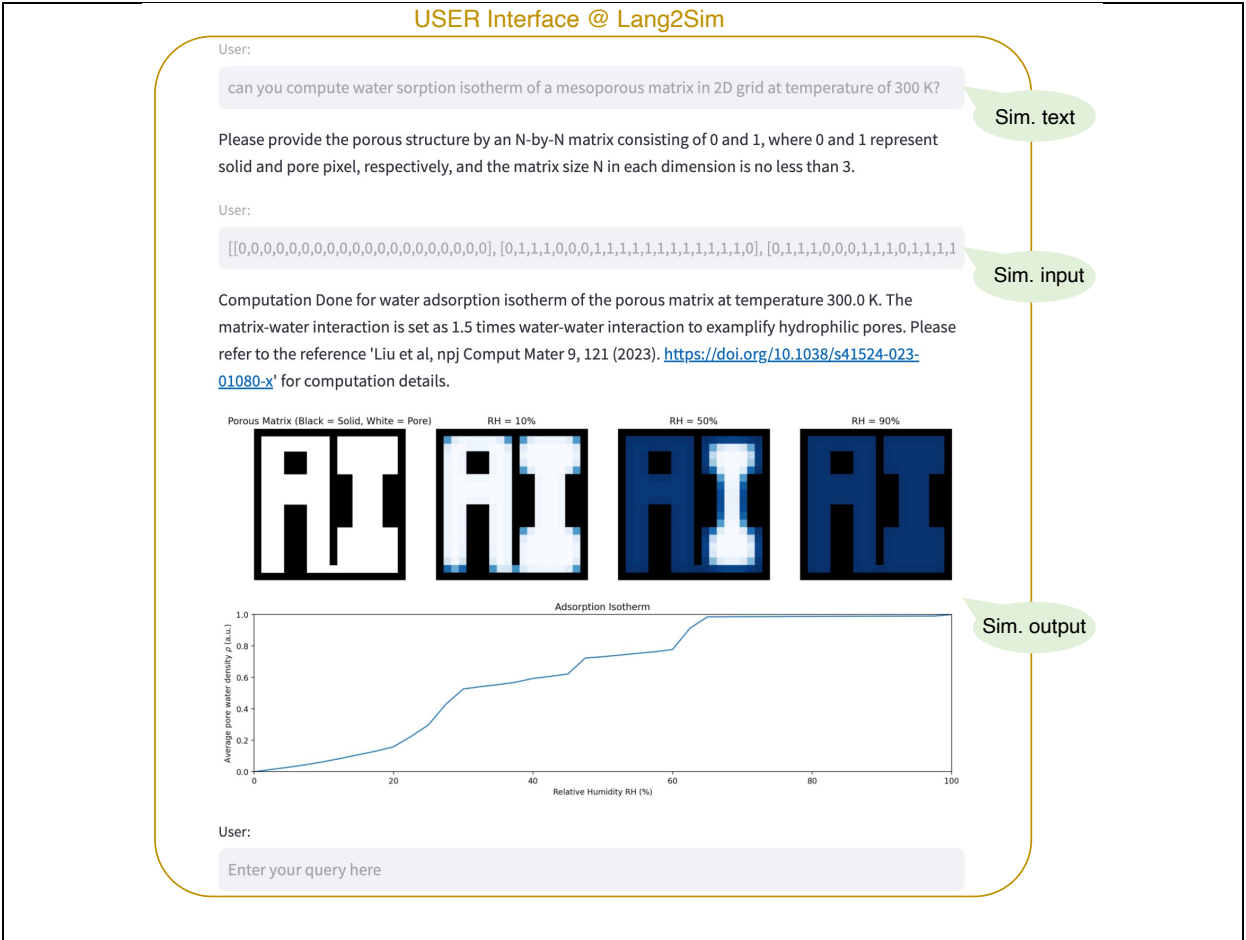

**Fig. 3: Lang2Sim example run on User-Interface (UI).** When inputting a simulation text, one of the LM chains on Lang2Sim platform (see Fig. 2b) is activated to interact with UI-detected text and to target its chain-end simulator, as exemplified herein by the scenario of water sorption simulation [30–32].

## 2.2 Simulation-Type module rationalizing the categorization of simulators

In the following sections, we take a closer inspection into the chain activation mechanism in each slice of the three-module platform. Firstly, we investigate the Simulation-Type module built by the assembly of LM-Type agents, wherein the module functionality is inherited from its LM-Type agent-functionality.

Figure 4a shows the module architecture built by chains of LM-Type. Along one of the chains, LM-Type outputs a simulation type within the categorization of its preceding LM-Type output, so as to empower the module to rationalize the categorization of a simulation type in human-language space. Note that, when the UI-detected text is irrelevant to the agent actions in categorizing simulations, LM-Type agent is designed to generate a type-hint as human-language guidance for the agent activation and, simultaneously, filter out these irrelevant texts to keep the textual description succinct.

Figure 4b provides a categorization hierarchy of materials simulations based on a sequential, classified description of simulation scale, functionality, and toolkit, by taking the example of 2D-LDFT sorption simulation [30–32]. Unlike the casual nature of untempelated human language [50], the templatized hierarchy reshapes the human-language space into concise representations, thus greatly reducing the space dimensionality and simplifying the sim–lang landscape thereof (see Fig. 4b). More details about the module architecture are provided in Methods section. Notably, this landscape simplification is key to facilitate the precise Lang2Sim conversion by naively activating the target action of LM-Type agent [25,26], regardless of the diversity of UI-detected texts.

Next, we visualize a chain of agent-action activation in the module by, along the chain, triggering the type-hint of each LM-Type to display on User-Interface (UI). Figure 4c shows an example of the module interplay with UI in languaging a 2D-LDFT sorption simulation [30–32], where the input text is set on purpose to mismatch the present LM-Type actions directed to their subtype LM-agents, so as to trigger a sequence of type-hint displays on UI. Accordingly, the textual description is refined at each response-step to approach the templatized language-features responsible for the present agent activation [25]. By iterating the description refinement step, an agent-action chain is activated throughout the module to reach its chain-end Simulator-Action module (see Sec. 2.3), i.e., 2D-LDFT simulation module herein [30–32]. It should be pointed out that, despite its intelligent activation of agent actions, the module agent hierarchy may deserve more fine-tuning [46–48], along with some complementary delicate formatting (if any) to simplify the sim–lang landscape to more pronounced extent, thus boosting its navigation intelligence.

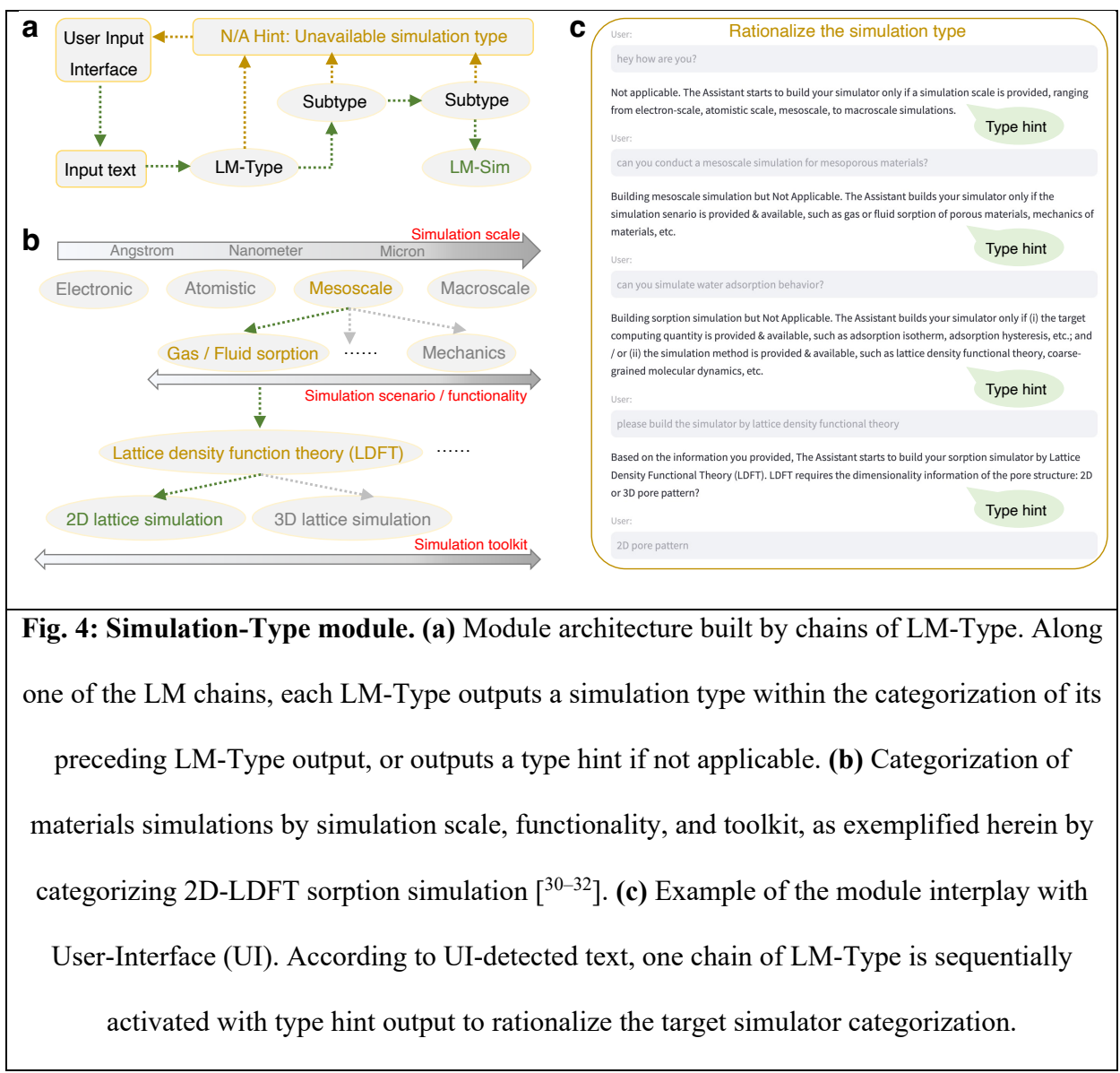

**Fig. 4: Simulation-Type module. (a)** Module architecture built by chains of LM-Type. Along one of the LM chains, each LM-Type outputs a simulation type within the categorization of its preceding LM-Type output, or outputs a type hint if not applicable. **(b)** Categorization of materials simulations by simulation scale, functionality, and toolkit, as exemplified herein by categorizing 2D-LDFT sorption simulation [30–32]. **(c)** Example of the module interplay with User-Interface (UI). According to UI-detected text, one chain of LM-Type is sequentially activated with type hint output to rationalize the target simulator categorization.

## 2.3 Simulator-Action module customizing the simulator input–output combination

Lying at the chain end of LM-Type agents from Simulation-Type module (see Fig. 4a), we now examine the Simulator-Action module built by an LM-Sim agent that contains multiple simulator actions, as illustrated in Fig 5a. Similar to its preceding module, the LM-Sim agent is activated to take an action targeting a simulator matching UI-detected text, or generate an action-hint of simulator availability and filter out the present chat-history texts if irrelevant. However, unlike LM-Type outputting a simulation type, LM-Sim directly outputs the target-simulator action, as derived from the last simulation type that ends the subtype categorization (see Fig. 4a and 5a). Note that, the ending subtype is regulated as an indivisible "element tool", e.g., 2D-LDFT simulation herein [30–32], with enclosed both an invariant tool function and its variant input–output pair (see Fig. 5b). Based on this regulation, each simulator within the subtype is defined and interpreted as a customization of the input and output variants in the same tool function.

Figure 5b shows the customization of multiple simulator-variants sharing the same 2D-LDFT tool function, by on-demand combination of its input–output variants, including the input parameters of porous matrix, temperature, etc., and the output quantities such as adsorption isotherm and hysteresis [30–32]. Details of 2D-LDFT tool function are provided in the Methods section. In the same spirit, by customizing its input–output pair, each "element tool" cultivates multiple simulators sharing the same tool function but diversifying in their functionalities. Figure 5c shows two examples of the module interplay with UI to evoke two different 2D-LDFT simulators that compute water adsorption hysteresis and isotherm, respectively, by providing the textual description of computing quantities targeted by the simulator output. Notably, from the language viewpoint, interpreting each simulator of the same subtype by its input–output pair can significantly reduce the dimensionality of human-language space and, therefore, simplify the sim–lang landscape to facilitate the precise transform of textual description into the target simulation engine.

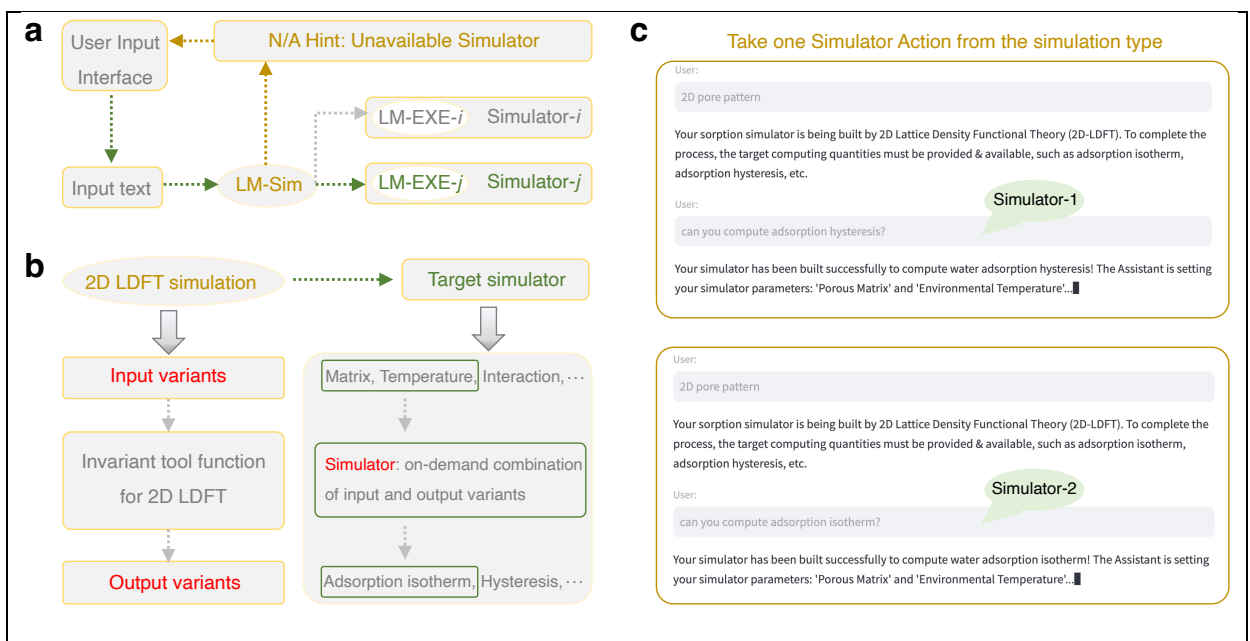

**Fig. 5: Simulator-Action module. (a)** Module architecture built by an LM-Sim that contains multiple simulators. Along one chain of LM-Type in Simulation-Type module (see Fig. 4), LM-Sim lies at the chain end to activate the simulation type including multiple simulator-variants and outputs a simulator action, or outputs an action hint if not applicable. **(b)** Customization of simulator-variants within one simulation type—i.e., simulators sharing an invariant tool function, by on-demand combination of its input and output variants, as exemplified herein by customizing 2D-LDFT sorption simulation [30–32]. **(c)** Examples of the module interplay with User-Interface (UI). According to UI-detected text, different 2D-LDFT simulator variants are evoked to compute water adsorption hysteresis (Simulator-1) and isotherm (Simulator-2), respectively.

## 2.4 Simulator-Execution module transforming textual input to simulator input

Finally, after evoking the target simulator from Simulator-Action module (see Fig. 5a), we investigate the workflow to execute the simulator residing in Simulator-Execution module. Figure 6a shows the module architecture built by an (or a stack of) LM-EXE linked to the simulator tool function, wherein the LM-EXE is designed to distill simulator inputs from UI-detected text and transform them into simulator-executable formats. In scenarios requiring multiple-format inputs, e.g., number versus matrix, a stack of LM-EXE agents is designed with each agent taking charge of one input-format and, upon activation, formatting its textual input into simulator-executable input. If encountering an irrelevant text, the agent generates an input-hint as the format guide and filters out the present chat-history texts for conciseness. More details about LM-EXE agents are provided in the Methods section.

Figure 6b illustrates a stack of two LM-EXE agents that transform textual description into two executable inputs of a 2D-LDFT simulator, wherein the simulator requires a matrix-format input of 2D porous pattern and a number-format input of environmental temperature, so as to compute water adsorption isotherm for any 2D porous matrix at different temperatures [30–32]. Accordingly, the two agents, referred to as LM-EXE-Matrix and LM-EXE-Number, are responsible for outputting a float 2D-matrix and a float number, respectively. To visualize the two agent-functionalities, Figure 6c provides an example of the module interplay with UI. By setting intentionally the absence of simulator inputs, the initial UI-detected text is sequentially fed into LM-EXE-Number and -Matrix to activate their input-hint guides, respectively. According to each agent hint, new textual input typed on UI is directly fed into the agent and, once the two agents complete the formatting of their own textual inputs, the simulator tool function is equipped to execute with these inputs and, eventually, outputs the target computing quantities. Once finished, the simulator clears all previous chat-history to release memory and, as a compensation, adds into the chat-history a summary of the present simulator settings to best balance the memory limit and information completeness [11,51] (see Sec. 2.5).

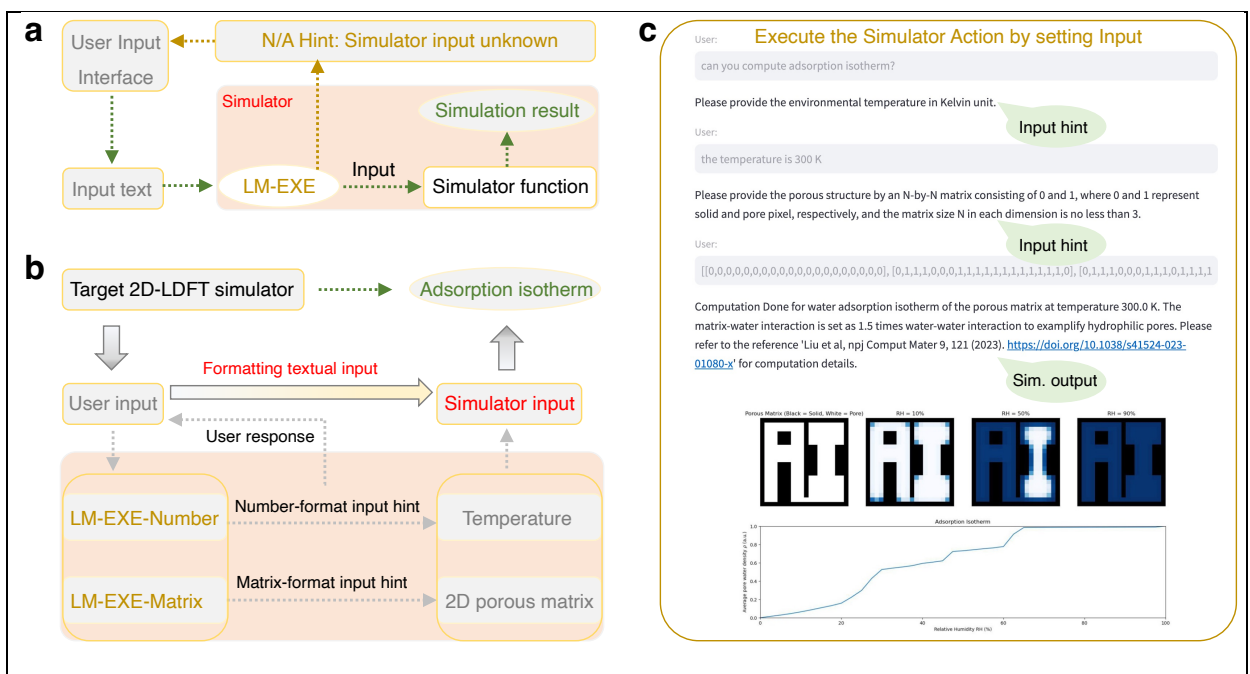

**Fig. 6: Simulator-Execution module. (a)** Simulator architecture built by a stack of LM-EXE linked to a tool function. When a simulator is evoked from Simulator-Action module (see Fig. 5), each LM-EXE of its stack takes charge of a simulator input format (e.g., number or matrix) and outputs the formatted input, or outputs an input-hint if not applicable. **(b)** Transform of textual input into simulator-executable input by functionalizing a stack of output-formatted LM-EXE, as exemplified herein by LM-EXE-Number and -Matrix in charge of, respectively, number- and matrix-format output fed into a target 2D-LDFT sorption simulator [30–32]. **(c)** Example of the module interplay with User-Interface (UI). According to UI-detected text, the stack of LM-EXE is sequentially activated with an input-hint to execute the target simulator.

It is worth mentioning that, to realize its functionality, each LM-EXE agent is built as a combination model consisting of two subagents, namely, LLM-Action and LM-Pattern relying on the techniques of language transformer [47] and similarity search [49], respectively. When the initial UI-detected text comes to LM-EXE, the LLM-Action subagent is activated to output an action of the simulator-input state by "known" or "unknown", and subsequently, the input state is passed to the LM-Pattern subagent for format-validation or -transformation if the simulator input is known, or for input-hint if unknown or invalid. Once the input-hint is displayed on UI, new textual input typed on UI would keep a direct feed into LM-Pattern until the input-formatting completion. Unlike the LM-agents in two preceding modules that solely rely on LLM technique [47]—the same as LLM-Action, the incorporation of LM-Pattern allows a supplementary validation or modification of the LLM output, and more necessarily, enables matrix formatting that is out of the present LLM capacity [52]. Overall, it concludes that LM-agents can be constructed toward enhanced intelligence and reliability, by leveraging both the long-text predictability of language transformer and the patternization capacity of similarity search model [49].

## 2.5 History chat effect on Lang2Sim modules

As a final critical examination to the intelligence of Lang2Sim platform, we investigate herein the history chat effect on each of the three platform-modules. In an ideal scenario, we expect that the chat-history memory contains all essential text tokens to target an LM-agent action and, at the same time, filters out all irrelevant texts to prevent the chat memory from exceeding its maximum token limit set by the LLM agent [11,48]. Namely, the memory setting of chat-history needs to balance both its memory limit and information completeness [11,51], so that each LM-agent appears intelligent to takes an action based on not only the present textual input, but also the information stored in chat-history memory.

Figure 7 shows an example of chat history effect on languaging a succeeding simulator after the completion of a simulator execution, by taking the example of 2D-LDFT sorption simulation [30–32]. Upon the simulator outputs displayed, UI terminates its interplay with Simulator-Execution module and, in turn, takes one step back redirected to Simulator-Action module. As mentioned earlier, upon simulation completion, the chat-history memory is reconfigured as a memory-note summarizing the present simulator settings (see Sec. 2.4). This memory-note, together with new UI-detected text, is delivered to the LM-Sim agent for next simulator-action and, when entering into the target simulator, would activate LM-EXE—more specifically, the LLM-Action subagent—to extract the simulator-input state. As such, the last simulator settings are intentionally parsed to the present modules, including both the last simulator-action and its inputs, so that the Lang2sim platform enables an uninterrupted execution of (i) new simulator-action whose input settings inherited from the historical input (see left panel of Fig. 7) or (ii) new simulator-input whose simulator-action inherited from the historical action (see right panel of Fig. 7).

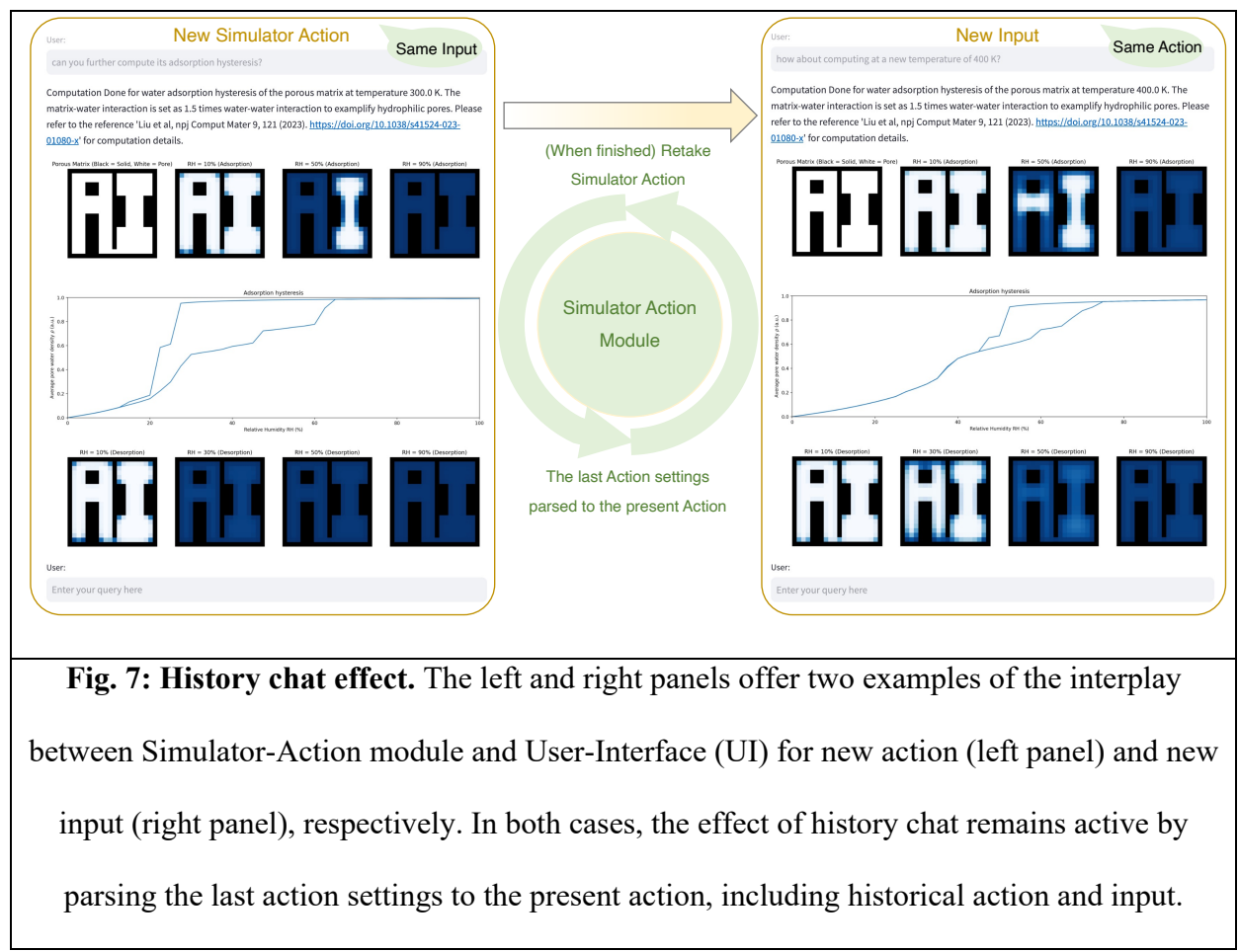

**Fig. 7: History chat effect.** The left and right panels offer two examples of the interplay between Simulator-Action module and User-Interface (UI) for new action (left panel) and new input (right panel), respectively. In both cases, the effect of history chat remains active by parsing the last action settings to the present action, including historical action and input.

Note that, to optimize the chat-history, Lang2sim modules accommodate chat-history processing in accordance to their agent functionalities. For LM-Type agents categorizing simulation types, since their navigation hierarchy contains by itself a history of action chain for each agent located in the hierarchical map, new textual input typed on UI is the only chat-history essential for the agent activation. Similarly, for an LM-Sim agent customizing simulators of the same subtype, its chat-history is inherited from its preceding LM-Type agent at its first activation and, afterwards, the agent is solely activated by new textual input typed on UI, or as discussed above, together with a memory-note of the last simulator-action settings if exists (see Fig. 7). Finally, for an LM-EXE agent distilling simulator input with two subagents, its chat-history is inherited from its preceding LM-Sim agent at its first activation and, afterwards, the LLM-Action subagent goes to dormancy until simulation completion, with the LM-Pattern subagent solely activated by new textual input typed on UI if there remains absent simulator-input. More details about chat-history processing are provided in the Methods section. Overall, by leveraging the distinct processing of chat-history for different agents, we demonstrate that the Lang2Sim platform exhibits pronounced intelligence to grasp the human-language organization featuring a significant extent of unstructured disorder.

## 3. Conclusions and Outlook

Overall, this work establishes an Lang2Sim platform built by three-module hierarchical architecture, with each module empowered by a functionalized assembly of LM-agents. Notably, the platform enables precise transform of human language to simulation engine in the complex sim–lang landscape, as exemplified herein by languaging a 2D-LDFT sorption simulation [30–32]. Essentially, this transform intelligence is enabled by simplifying the sim–lang landscape, including (i) rationalizing the categorization of simulation type, (ii) customizing the simulator input–output combination, and (iii) distilling the simulator input into executable format. Importantly, relying on distinct chat-history processing for each LM-agent type, the platform balances the memory limit of chat-history and its information completeness, so as to intelligently interplay with new UI-detected text by incorporating the key messages stored in chat-history memory.

This work unveils the potentiality of language models as an intelligent platform to unlock the era of "languaging a simulation engine". We expect that the platform introduced herein would modestly stimulate new developments in that direction. As future opportunities, we envision that the functionalities of Lang2Sim platform would be empowered by advancing several branch directions, as illustrated in Fig. 8, including (i) templatized accommodation of miscellaneous simulators [5,6], (ii) mutual integration of simulations and machine learning [6,7], and (iii) tabulated schedule of all-in-one model execution [42]. Their unification within one framework would present an extreme thrust to extend the capacity of Lang2Sim functionality, wherein the boundary between simulation, machine learning, and human language would eventually fade.

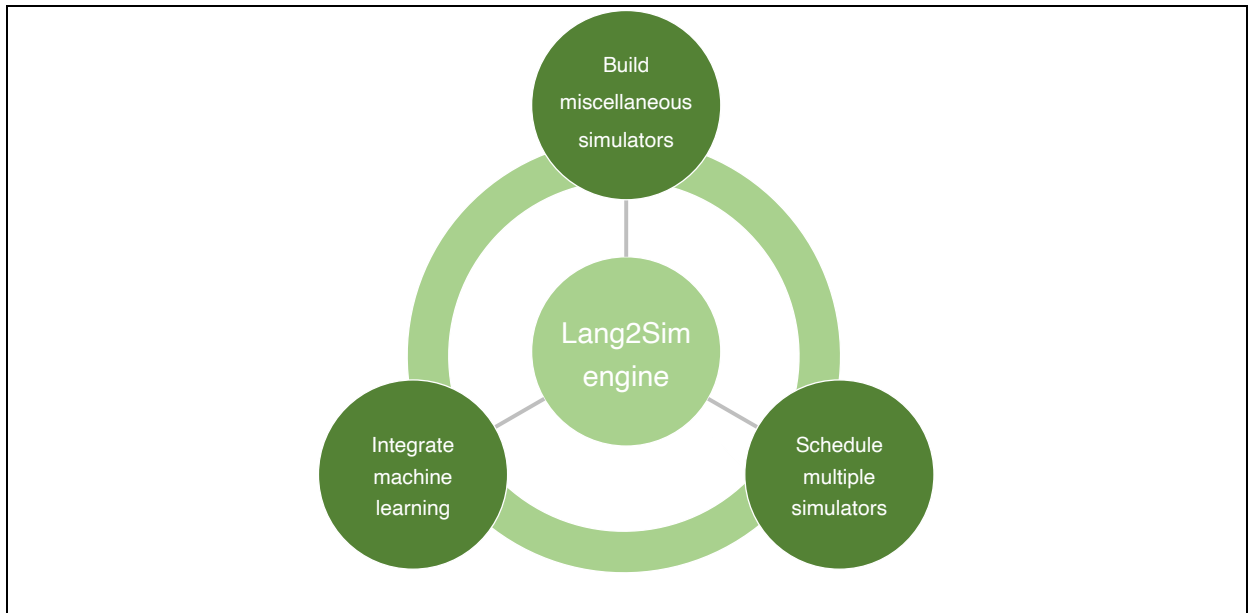

**Fig. 8: Future development of Lang2Sim platform.** Built upon the three-module hierarchical architecture (see Fig. 2), the functionality of Lang2Sim platform would be empowered by (i) templatized accommodation of miscellaneous simulators [5,6], (ii) mutual integration of simulations and machine learning [6,7], and (iii) tabulated schedule of all-in-one model execution [42].

## 4. Methods

### 4.1 Sorption simulation by lattice density function theory (LDFT)

Relying on 2D lattice density functional theory (2D-LDFT) [30–32], we construct two simulators that compute water adsorption isotherm and hysteresis, respectively, for any 2D porous matrices $M$ at different temperatures $T$. Other parameters are fixed based on Ref. [30]. Both simulators share an invariant tool function $f$ described by 2D-LDFT formalism [30–32], that is, the water density $\rho_i$ at pixel $i$ of a 2D grid under a relative humidity (RH) is given by:

$$\rho_i = f(M, T, \text{RH}) \qquad \text{Eq. (1)}$$

where $\rho_i$ ranges from 0 to 1 to represent that the pore is fully empty or saturated with water, respectively. Then the average pore water density $\rho$ is calculated by averaging $\rho_i$ over all pore pixels. The simulator built to compute adsorption isotherm is to apply Eq. (1) at RH = 0-to-100% with an increment dRH (here, dRH = 2.5%). Similarly, the simulator built to compute adsorption hysteresis is to further flip the RH values and apply Eq. (1) to obtain the desorption isotherm. More details of LDFT formalism can be found in Ref. [30–32].

### 4.2 Three functionalized language models (LM-agents)

We then construct three types of LM-agents, i.e., LM-Type, LM-Sim, and LM-EXE. Note that LM-EXE consists of two subagents, viz., LLM-Action and LM-Pattern. Among all the agents, LM-Type, LM-Sim, and LLM-Action are built by a prompt-engineered and fine-tuned LLM from LLaMA2-7B-chat [46–48]. The prompt templates for the three agent-types are prescribed to target a json-format action of simulation type, simulator index, and simulator-input state, respectively [24–26]. We then prepare a training set of ~20 input–output text pairs per agent for fine tuning by Low-Rank Adaptation (LoRA) [47,53], wherein the input is formatted by its prompt template and, accordingly, the output is the target json-format action. Unlike LLM-based agents [24–26], LM-Pattern is built by a similarity search model using regular expression patterns [49]. Specially, we construct two types of LM-Pattern agents, i.e., LM-Pattern-Number and -Matrix that adopt a float-number pattern and a float 2D-matrix pattern, respectively, so as to format the textual inputs of 2D porous matrix and environmental temperature for 2D-LDFT simulator [30–32]. Note that, if encountering irrelevant textual input, all LM-agents are designed to take their hint action for guidance.

### 4.3 Architecture of Lang2Sim modules

Relying on the three LM-agent types, we now build the three modules of Lang2Sim platform (see Fig. 2b), including Simulation-Type, Simulator-Action, and Simulator-Execution module, built by an assembly of LM-Type, LM-Sim, and LM-EXE, respectively. The Simulation-Type module assembles LM-Type agents as a hierarchy based on simulation scale, functionality, and toolkit. Taking 2D-LDFT simulation for instance [30–32] (see Fig. 4b), as one branch of agent hierarchy in this module, the first-level agent LM-Scale takes charge of categorizing simulation scale into four subtypes as the second-level agents, namely, LM-electronic, -atomistic, -mesoscale, and -macroscale. By activating LM-mesoscale agent, one of the third-level agents is about to select according to their simulation scenarios or functionalities, such as LM-sorption and -mechanics, with LM-sorption matched herein. Then based on the target computing quantities, one candidate computational method or toolkit is recommended in the fourth-level agents, including LM-LDFT or -CGMD short for coarse-grained molecular dynamic [54,55]. Finally, the LM-LDFT is further divided into two subtypes, i.e., LM-2D-LDFT and -3D-LDFT as the ending-level agents [30–32].

Here, we take the LM-2D-LDFT agent as LM-Sim in Simulator-Action module under investigation. Accordingly, we build two simulators to be incorporated in the module, namely, 2D-LDFT-isotherm and -hysteresis that compute water adsorption isotherm and hysteresis, respectively [30–32] (see Sec. 4.1). By outputting one simulator action, e.g., 2D-LDFT-isotherm, the agent activates this simulator as Simulator-Execution module. Inside the simulator, textual input goes through LLM-Action and LM-Pattern to transform into a float number for temperature input and a float 2D-matrix for porous pattern input. Finally, the executable inputs are fed into the simulator tool function to output the target computing quantities such as water adsorption isotherm [30–32].

### 4.4 Interplay between Lang2Sim modules and User interface (UI)

Next, we construct a User-Interface (UI) to visualize the interplay between the three modules and textual inputs. The UI is essentially a text-box built to receive textual input and display an output response from the modules. Besides detecting textual input, the UI is used to link the text-box to a specific LM-agent that is being activated at present. Even after closing and reopening UI, this linkage is designed to remain valid until the next LM-agent on the chain is activated. Once a simulation is finished, the linkage between UI and the Simulator-Execution module is terminated, with an "one-step-back" old linkage reestablished between UI and the LM-Sim agent in the preceding Simulator-Action module. Moreover, to fully visualize the chat-mode interplay, the UI is designed to display all chat records between user inputs and module outputs, which remains valid even after closing and reopening UI, by loading all the records back on screen. Note that, although all chat records are displayed on UI, each LM-agent has a distinct processing of chat-history that does not necessarily memorize all the chat records (see Sec. 4.5).

### 4.5 Window selection and processing of history chat

Finally, we optimize the chat-history memory by distinctly selecting and processing the chat records for each LM-agent type, including LM-Type, LM-Sim, LLM-Action, and LM-Pattern. Among these agents, LM-Type is designed to activate once and to interplay solely with the present UI-detected text input, that is, without history memory needed. LM-Sim adopts the same chat-history setting at its first activation, but once a simulation is finished, the chat-history memory is cleared to empty and, in turn, is added with a summary note of the simulation settings. LLM-Action shares the chat-history fed into LM-Sim, same for LM-Pattern at its first activation. However, once a simulator-input hint is displayed on UI, LM-Pattern is designed to solely interplay with new textual input typed on UI, with no history chat fed into the agent. To a pronounced extent, these distinct chat-history settings significantly balance its memory limit and information completeness [11,51].

# Acknowledgements

**Funding:** H.L. acknowledges funding from the Fundamental Research Funds for the Central Universities under the Grant No. YJ202271, and the National Natural Science Foundation of China under the Grant No. 52303042. **Author Contributions:** Conceptualization: HL; Methodology: HL; Investigation: HL and LL; Visualization: HL and LL; Supervision: HL; Writing (original draft): HL; Writing (review and editing): HL and LL. **Competing interests:** The authors declare that they have no competing interests. **Data and materials availability:** All data needed to evaluate the conclusions of this study are present in the paper and available from Dr. Han Liu upon reasonable request.